\documentclass[twocolumn]{article}

\usepackage[utf8]{inputenc}
\usepackage{booktabs}
\usepackage{libertine}
\usepackage[libertine]{newtxmath}
\usepackage{amsmath}
\usepackage[top=2.5cm, bottom=2.5cm, left=2.5cm, right=2.5cm]{geometry}
\usepackage{hyperref}
\usepackage{enumitem}
\usepackage{ulem} \usepackage{xcolor}
\usepackage{graphicx}
\usepackage{caption}
\usepackage{subcaption}

\definecolor{ForestGreen}{rgb}{0.13, 0.55, 0.13}
\newcommand{\op}{\textrm{op}} \newcommand{\model}{\mathcal{M}} 
\title{Zoetrope Genetic Programming for Regression}
\author{Aurélie Boisbunon$^*$, Carlo Fanara$^*$, Ingrid Grenet$^*$,\\ Jonathan Daeden$^*$, Alexis Vighi$^*$, Marc Schoenauer$^\dag$}

\date{\normalsize $^*$MyDataModels, Sophia Antipolis, France\\
$^\dag$INRIA Saclay, LRI, Orsay, France}

\begin{document}

\maketitle

\begin{abstract} 
The Zoetrope Genetic Programming (ZGP) algorithm is based on an original representation for mathematical expressions, targeting evolutionary symbolic regression. The zoetropic representation uses repeated fusion operations between partial expressions, starting from the terminal set. Repeated fusions within an individual gradually generate more complex expressions, ending up in what can be viewed as new features. These features are then linearly combined to best fit the training data. ZGP individuals then undergo specific crossover and mutation operators, and selection takes place between parents and offspring. ZGP is validated using a large number of public domain regression datasets, and compared to other symbolic regression algorithms, as well as to traditional machine learning algorithms. ZGP reaches state-of-the-art performance with respect to both types of algorithms, and demonstrates a low computational time compared to other symbolic regression approaches.
\end{abstract}

\section{Introduction} 

Symbolic Regression (SR) is a supervised learning approach
that consists in searching through a vast space of predictive models. This space encompasses the rigid linear and polynomial models by enabling other transformations such as trigonometric and logarithmic, as well as the generalized additive models by allowing (linear and) nonlinear combinations of the transformed variables  (see \cite{vzegklitz2017symbolic} and references therein). SR models are often represented via expression trees, and 
may include decision tree models if we consider equality and inequality operators instead of functions (see e.g. the Boolean multiplexer in \cite{Koza92}). Models can also be unravelled through mathematical formulae, which make them much more interpretable than other tree or network-based machine learning algorithms such as random forests \cite{breiman2001random} or neural networks
  where, with few recent exceptions \cite{kim2020integration}, the relationships among the variables remain hidden.  
  SR thus offers a good tradeoff between flexibility and interpretability. Moreover, it does not need large numbers of observations as in deep neural networks, and can be applied to smaller datasets (typically from several dozens to tens of thousands). 

The history of SR is closely related to that of Genetic Programming (GP), starting with the early works of Koza \cite{Koza92,langdon2013foundations}. Indeed, most of the approaches for SR are GP algorithms (often denoted GPSR), as these offer a nice framework with expression trees representing the potential models. GPSR algorithms start with a pool of initial models, which are then iteratively and randomly perturbed to create new ones, until the one that fits the data best is finally selected. Variants to this scheme are discussed in \cite {vanneschi2014survey}, whereas newer approaches are enlisted in \cite{virgolin2020improving}.
On the contrary, "standard" machine learning algorithms are not well suited for the complex task of optimizing both parameters and model shape at the same time \cite{poli2008field,vries2018sensitivity}. Several exceptions are worth noting, combining the idea of expression trees with classical ML \cite{mcconaghy2011ffx} or with neural networks \cite{kim2020integration}.

While SR is very present in the field of Evolutionary Algorithms (EA) \cite{mcconaghy2011ffx}, it is almost completely absent in Machine Learning (ML) reference books \cite{hastie2009elements,bishop2006pattern} and toolboxes \cite{scikit-learn}. 
Possible reasons could be the following: first, there are many SR algorithms in the literature, each offering various advantages \cite{vries2018sensitivity}, and it might be difficult to know which one to use and how to tune their often large number of parameters; second, these algorithms are often slower than most ML algorithms, with performance that did not match those of e.g. random forests up until recently; finally, earlier works mostly tested symbolic regression on synthetic data with a known equation involving few input variables, with the aim of recovering exactly this equation \cite{korns2011accuracy}, and not often on real datasets with more than 5 variables. For a critical view on SR benchmarks we refer the reader to \cite{mcdermott2012genetic,olson2017pmlb} and references therein.

These limitations have been overcome in the last decade, at least partially. The issue of computational time has been treated by Geometric Semantic Genetic Programming (GSGP) 2.0 \cite{castelli2019gsgp} with the proposition of a very efficient algorithm. However, this efficiency comes at the cost of interpretability, as the use of geometric semantic variation operators results in exponentially growing trees. The performance of GPSR has been increased for instance by the combination of GP with more standard ML approaches \cite{arnaldo2014multiple}. Finally, novel benchmarks were established lately that also compare SR and classical ML algorithms on real datasets~\cite{orzechowski2018we,vzegklitz2020benchmarking,affenzeller2019white}. These benchmarks show that the performance of random forests can be matched by increasing the number of individuals and generations, considerably slowing down the computations. So the issue remains for GPSR to get good performance in a reasonable time without losing its characteristic interpretability. A recent and promising work has been proposed in that sense \cite{lacava2019learning}, as well as our own work that we present here.

In this article we present a new GP algorithm called Zoetrope Genetic Programming (ZGP), which brings the following main contributions: 
(1) a new and unseen representation of models, allowing fast computation {and} feature engineering, {while keeping the interpretability advantage of most SR methods} { through the explicit model formula;} 
(2) novel mutation and crossover processes, leading to improvement of models over the generations; (3) performance that is comparable to the best ML (Gradient Boosting) and SR algorithms.
    While ZGP can handle the three main supervised learning tasks, namely regression, binary classification and multiclass classification, we focus here only on the regression one. 

The paper is organized as follows. 
Section~\ref{sec:background} briefly presents the general context and introduces GPSR state-of-the-art frameworks which are related to our work. Section~\ref{sec:zgp} describes the entire ZGP algorithm, with its uncommon representation of individuals and variation operators, as well as the choices for fitness and cost.
Section~\ref{sec:applications} presents {and discusses} the results of our benchmark against state-of-the-art SR frameworks and ML algorithms on 98 regression datasets.
Finally, we {conclude on our main contributions in  Section~\ref{sec:discussion}, and suggest potential future work.}

\section{Background}\label{sec:background}

\subsection{Context}

Given a dataset $\mathcal D$ made of i.i.d. observations 
$(X_i,y_i) \in \mathbb{R}^d \times \mathbb{R}$ with $X_i = (X_{i1}, \ldots,  X_{id})$,
the goal of regression algorithms is to find a function $\model:\mathbb{R}^d\mapsto \mathbb{R}$ modelling the link between $y$ and $X$, such that it generalizes well on unseen data from the same distribution.

As common in regression problems, as performance measure for $\model$ on dataset $\mathcal D$ we use the Mean Squared Error (MSE) defined by:
\begin{equation}
    MSE(\model, {\mathcal D}) = \frac{1}{\#{\mathcal D}} \sum_{(X_i, y_i)\in {\mathcal D}}\left(y_i-\model(X_i)\right)^2 .
\end{equation}

Any dataset $\mathcal D$ used in this work will be divided into training ($\mathcal D_T$), validation ($\mathcal D_V$) and test, or holdout ($\mathcal D_H$) sets. The holdout set $\mathcal D_H$ is never to be seen during the learning procedure, and is only used to assess the final performance of the model. The uses of the training and validation sets is detailed in Section \ref{sec:algo}.

\subsection{Related work}

As mentioned above, several SR frameworks have been recently proposed and already compared to classical ML algorithms. First, some SR techniques are not based on evolutionary process. For example, Fast Function Extraction (FFX)~\cite{mcconaghy2011ffx} only generates a large set of linear and non linear features and then fits a linear model on the features using elastic net \cite{zou2005regularization}. While the deterministic part can be attractive to avoid getting different models from one run to another, it turns out that FFX often results in much larger models than conventional GP. {Evolutionary Feature Synthesis (EFS)~\cite{arnaldo2015building} uses a similar idea, but avoids building the basis entirely by randomly generating them. It is however not a GP algorithm. }
The idea of linearly combining branches of a tree is also very present in GP, {as it allows the construction of new features.} Multiple Regression Genetic Programming (MRGP)~\cite{arnaldo2014multiple} combines all the possible subtrees of a tree through LASSO~\cite{tibshirani2011regression}, thereby decoupling the linear regression from the construction of a tree. More recently, La Cava et al. 
developed Feature Engineering Automation Tool (FEAT)~\cite{lacava2019learning}, 
which trades conciseness for accuracy. It is a stochastic optimization providing a succinct syntactic representation with variable dependencies explicitly shown (in contrast to the semantic approach \cite{ribeiro2016should}). {Another related recent work is the Interaction-Transformation Evolutionary Algorithm (ITEA)~\cite{defranca2020interaction}, which builds generalized additive models including interactions between variables.}

{The efficiency of GP has been another direction of study.}
Geometric Semantic Genetic Programming (GSGP) \cite{moraglio2012geometric} is a technique combining trees to get new individuals and adds semantic methods for crossover and mutation in order to introduce a degree of 'awareness'. However, in GSGP the generated individuals are larger than their parents, resulting in large bloat, and longer computing times. This is addressed by using a practical development environment, GSGP-C++ \cite{castelli2019gsgp} with operators in native C++. Finally, other frameworks propose efficient selection techniques. Age-Fitness Pareto Optimization (AFP) \cite{schmidt2011age} is meant to prevent premature convergence in evolutionary algorithms by including age as an optimization criterion using a Pareto front between age and fitness. This allows younger individuals to compete with older and fitter ones. Also, $\epsilon$-lexicase selection (EPLEX)~\cite{lacava2016epsilon} performs parent selection according to their fitness on few random training examples, dropping all the population individuals with error higher than the best error. {This selection technique is used in FEAT.}

{With respect to the above works, ZGP proposes two novelties. First, ZGP uses a parametric representation for its models. Second, within its complex genotype-to-phenotype mapping, ZGP borrows to Geometric Semantic Crossover \cite{moraglio2012geometric}, and thus compensates the could-be limitations of a fixed representation by creating a richer set of smoother trajectories in the space of all possible analytical expressions / programs. Furthermore, this process sets a strict bound on the complexity of the resulting expressions, and thus limits the bloat.}

\section{The ZGP algorithm}
\label{sec:zgp}
The Zoetrope Genetic Programming\footnote{{ZGP is a proprietary algorithm with patent pending. An open source version is currently under development.}} (ZGP) algorithm {is based on the original {\em Zoetropic} representation for programs, together with the corresponding variation operators (crossover and mutation). However, the "natural selection" components of all evolutionary algorithms are here directly incorporated into the variation operators (i.e., selection takes place between the parents and their offspring only).} 
Furthermore, ZGP uses evolutionary components 
to build possible branches of a regression tree, and standard ML techniques to optimize the combination of those branches.

\subsection{The Zoetropic Representation}
\label{sec:z-forge}

\begin{figure}
    \centering
    \includegraphics[width=\linewidth]{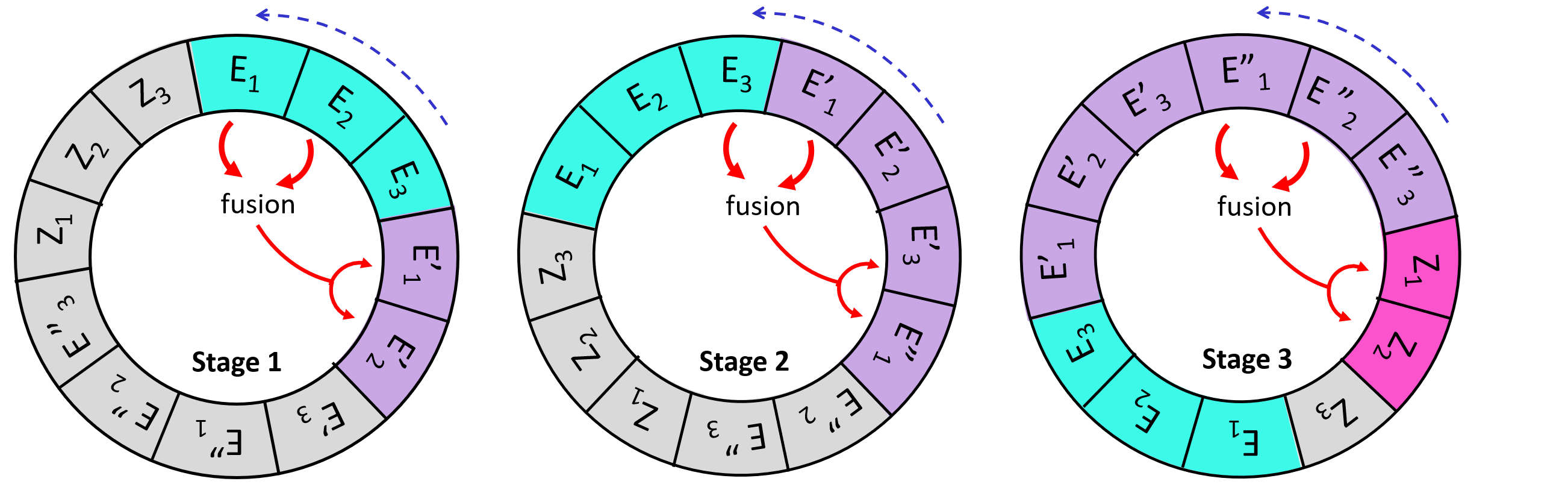}
    \caption{Illustration of Zoetropic representation building: for $n_e=n_m=3$, there are $n_f=4$ fusions in total, and for the sake of readability, the third one, generating $(E"_2, E"_3)$ from $(E'_2, E'_3)$, taking place between center and right figures, is not represented. Note that $Z_3=E"_3$ as no element is left for a fusion.} 
    \label{fig:stages_ZGP}
\end{figure}

This section describes both the genotype and the genotype-to-phenotype mapping  of ZGP individuals.
As in standard tree-based GP \cite{Koza92,banzhaf98}, ZGP individuals are built from a set of unary or binary operators $\mathcal O$ and a set of terminals $\mathcal T$,  variables of the problem and ephemeral constants. The genotype of a ZGP individual is built using  {\em elements} (partial expressions built on $\mathcal T$ and $\mathcal O$) and {\em fusion} operations (see below). Two parameters control the size of the genotype as well as the derivation of the corresponding phenotype (the final expression used to evaluate the fitness of the individual):  the number of initial elements $n_e$ and the number of {\em maturation} stages $n_m$. 
An individual is built as follows:

\smallskip  
\noindent{\bf Overview and notations} The elements used during the process can be seen as organized in $n_m$ levels, one per maturation step. The $n_e$ elements of level $k$, denoted $(E_1^k, \dots, E_{n_e}^k)$, are constructed by maturation step $k$ from the elements of level $k-1$. However, due to boundary conditions depending on the parity of $n_e$, it is more convenient to visualize all the elements in a circle (reminding the original zoetrope mechanism\footnote{The term zoetrope historically defines one of the first  animation devices before the camera, consisting of a cylinder with images inside, that  seem to be moving as the cylinder is turned.}). Figure \ref{fig:stages_ZGP} illustrates the creation process, but for the sake of simplicity, the elements of levels 0, 1, 2, 3 are denoted $E_i$, $E'_i$, $E''_i$ and $Z_i$ respectively (explanations below).

\smallskip  
\noindent{\bf Initialization} The $n_e$ elements $(E_1^0, \dots, E_{n_e}^0)$ are randomly drawn in $\mathcal T$, being a uniformly chosen variable with 90\% probability, or an ephemeral constant with 10\% probability. {The latter are uniformly drawn in $[C_{min}, C_{max}]$, for some user-defined parameters $C_{min}$ and $C_{max}$}. 

\smallskip  
\noindent{\bf Fusion} The fusion operation $\mathcal F$ transforms a pair $(E_i, E_j)$ of elements into a new pair $(E'_i, E'_j) = {\mathcal F}(E_i, E_j)$. It starts by computing
    \begin{equation}\label{eq:fusion}
        f(E_i, E_j) = r \cdot \op_1(E_i, E_j) + (1-r)\cdot \op_2(E_i, E_j),
    \end{equation}
    where $\op_i$, $i=1,2$ are operators uniformly chosen in $\mathcal O$, and $r=U[0,1]$ (in case $\op_1$ or $\op_2$ is unary, only $E_i$ is taken into account).
    Elements $E'_i$ and $E'_j$ are then defined by 
    \begin{equation*}
    \begin{split}
        E'_i & = b \cdot E_i + (1-b) \cdot f(E_i, E_j)\\
        E'_j & = (1-b) \cdot E_j + b \cdot f(E_i, E_j),
    \end{split}
    \end{equation*}
    where $b=U\{0,1\}$, i.e., one new element is equal to one randomly chosen original element, while the other is defined by Eq. (\ref{eq:fusion}). 
    The fusion $\mathcal F$ is defined by $(\op_1, \op_2, r, b)$. 
 
{Note that these fusion operations, and in particular Equation (\ref{eq:fusion}), are somehow similar to the Geometric Semantic Crossover \cite{moraglio2012geometric}. But the linear combination with random weight is done here at the level of simple operators, not subtree, and during the genotype-to-phenotype mapping, not during crossover. In both situation, this results in a smoother landscape than only allowing blunt choices between one or the other argument, offering more transitional states to the evolutionary process.}

\smallskip  
\noindent{\bf Maturation} 
    The $k^{th}$ maturation step, or stage, consists of the sequence of     $\lfloor{n_e/2}\rfloor$ fusions defining elements $E_i^{k}$ from pairs of elements $E_i^{k-1}$. 
    
\smallskip  
\noindent{\bf The Zoetrope model} After $n_m$ maturation steps, the $n_e$ elements of level $n_m$, called "Zoetropes", are linearly combined to obtain the final model (see Section \ref{linearRecombination}). 
    
\smallskip  
\noindent{\bf Complexity analysis} There are $n_f = n_m\cdot \lfloor{n_e/2}\rfloor+n_e\%2$ fusions in total. 
    At each fusion, the size of the elements increases by the application of Eq. (\ref{eq:fusion}). In terms of standard GP indicators (though we never express the ZGP models as trees), the depth of ${\mathcal F}(E_i, E_j)$ is three more than the maximum depth of $E_i$ and $E_j$. Hence the depth of the zoetropes is at most $3*n_m+1$.  The linear combination applied to the zoetropes using the $n_e$-ary addition operator can be viewed as adding two more levels of depth.         In particular, because all created individuals use the same template, the complexity of any ZGP model remains bounded. Therefore, ZGP individuals are not subject to uncontrolled bloat.

\subsection{Fitness and Cost Functions}
\label{sec:algo}

\subsubsection{Combination of zoetropes}\label{linearRecombination}
As said in previous Section, at the end of all fusions, the zoetropes are combined to obtain the full model as:
\begin{equation}\label{eq:model}
    \model_{\boldsymbol{\alpha}}(X) = \sum_{j=1}^{n_e} \alpha_j Z_j(X),
\end{equation}
for some weights $\boldsymbol{\alpha} = (\alpha_1, \ldots, \alpha_{n_e}) \in \mathbb{R}^{n_e}$.

\subsubsection{The Fitness Function}

The fitness function, used in Darwinian selection, is the MSE of the best linear combination of the zoetropes on the training set, $MSE(\model_{\boldsymbol{\alpha}^*}, {\mathcal D_T})$. In ZGP, this fitness function is applied within the variation operators, between parents and offspring. 
Furthermore, the best individual in the population w.r.t. the MSE on the validation set $\mathcal{D}_V$ is stored at every generation, and after the algorithm has stopped, the overall best of these best-per-generation is returned (still according to the MSE on the validation set).

\subsection{The variation operators}
\label{sec:ea-components}
This section introduces the representation-specific variation operators, i.e., crossover and mutation (the initialization has been described in Section \ref{sec:z-forge}). As said, in ZGP, the selection is made within these variation operators, between parents and their offspring, using the MSE for comparisons.
{Furthermore, the way these operators are applied is also specific. This section will hence describe the operators as well as the choice of the individuals they are applied to.}

\subsubsection{The Crossover Operator}
\label{sec:crossover}
The crossover process of ZGP uses two parents, but works one-way: it only  propagates components from the fittest parent to the  other one, somehow similarly to the InverOver operator for permutations \cite{tao1998inver}. It starts by selecting $n_t$ individuals uniformly from the population (as in standard tournament selection). Then, it randomly replaces some of the 'genes' of the weakest parent by the corresponding genes of the fittest parent. 
The genes to replace are randomly chosen from the initial elements (terminals in $\mathcal T$) and the fusions, each fusion being considered as a single gene here.

\subsubsection{Applying the Crossover} 
{Our GP strategy for applying the crossover amounts to repeat $\rho_X \cdot P$ times the above procedure (tournament of size $n_t$, one-way gift of genes from best to worst), for some hyperparameter} {$\rho_X \in [0,1]$. }
{Its actual implementation runs some tournaments in parallel (i.e., without replacement between the tournaments), in order to decrease the overall computational time. }

\subsubsection{Point Mutation}

The point mutation operator considers one parent, and works as expected: it replaces some 'genes' of the parents by random values. However, the fusions are here considered made of four 'genes' here, the four components $(\op_1, \op_2, r, b)$ (Section \ref{sec:z-forge}), that can be modified independently.
{For each point mutation, either one element or one fusion is randomly chosen from the "genes" and mutated.}

When an element is to be mutated, it is replaced by a constant with probability $\rho_{cst}$, or with a variable uniformly chosen (and different from the current one if the element is a variable). When replacing a variable with a constant, this constant is simply chosen uniformly in $[C_{min}, C_{max}]$. When mutating a constant $C$ to a new constant, an auxiliary constant $\hat{C}$ is uniformly drawn also in $[C_{min}, C_{max}]$, an operator $o$ is uniformly drawn in $\{\times, /, +, power, nil\}$, and $C$ is replaced by $C$ $o$ $\hat{C}$  (where $C nil \hat{C} = \hat{C}$).

When a fusion is to be mutated, only one (uniformly drawn) of its four components $\op_1, \op_2, b, r$ is modified. In case of an operator, a new operator is chosen uniformly in  $\mathcal{O}$. $r$ is modified by flipping one bit of its binary representation, and $b$ is simply flipped.
Note that in ZGP, each individual designated for mutation is actually mutated twice. A first point mutation is applied to a component (element or fusion) randomly chosen from the "effective components", i.e., the components which are actually used by the model, thus ensuring that the mutation has an impact on the model. A second point mutation is applied to a component randomly chosen from all the components (effective or not), allowing components free from fitness pressure to drift and preserve diversity once they become effective \cite{lacava2015genetic}.

\section{Experimental Validation}
\label{sec:applications}
This section describes and analyzes the performance of ZGP  on regression tasks with tabular data, and compares them with those of state-of-the-art symbolic regression and classical machine learning algorithms. 

\subsection{Experimental Setting}
The experiment closely follows the benchmark in \cite{orzechowski2018we}, where the algorithms were run on the Penn Machine Learning Benchmarks (PMLB) database \cite{olson2017pmlb}, a collection of real-world, synthetic and toy datasets, with a restriction to datasets with less than 3000 observations (small data regime).
We compare ZGP with the same SR algorithms as in \cite{orzechowski2018we}, namely MRGP \cite{arnaldo2014multiple}, GSGP \cite{moraglio2012geometric,castelli2019gsgp}, EPLEX \cite{lacava2016epsilon}, AFP \cite{schmidt2011age}, with the best parameters their authors found by 5-fold cross-validation. To this list, we also added the more recent FEAT \cite{lacava2019learning} and the deterministic FFX \cite{mcconaghy2011ffx}. We also chose those algorithms because of the availability of a Python interface (provided by the benchmark's authors in the case of GSGP and MRGP). Indeed, while many state-of-the-art SR algorithms are open source, their source code comes in different languages (C++, Java, Matlab), hence quite some work is needed to re-implement and  run those under the same conditions.
As for classical ML approaches, we chose the following algorithms from scikit-learn \cite{scikit-learn}: gradient boosting, random forests (RF), decision trees, elastic net, kernel ridge and linear SVR, the latter three being optimized by 5-fold cross validation. Finally, we added a multi-layer perceptron with keras \cite{chollet2015keras} as in our experience, the one from scikit-learn does not perform well in general. The parameters for each algorithm are provided in supplementary material.

The experiment consisted in 20 runs of each algorithm, based on the same splits of training and test sets (70-30\%) for all algorithms\footnote{Note that some of the algorithms, including ZGP, further split the training set into training and validation, the rate of which was let to each algorithm's default parameters.}. 
All datasets were standardized with scikit-learn' \textit{StandardScaler}. We computed the Normalized Root Mean Squared Error (NRMSE), i.e., the square-root of the MSE divided by the range of target values, the R2-score (computed with scikit-learn), and the computational time for each algorithm and each run. Note that in \cite{orzechowski2018we}, only 10 independent runs were run for each algorithm, with random train-test splits. However, given the variability of symbolic regression approaches, we believe it is more robust to increase the number of runs, and fairer to compare them on exactly the same data. 

All experiments were performed on a HP Z8 server with 40 cores\footnote{CPU Intel(R) Xeon(R) Silver 4114, 2.20GHz, 64 GigaBytes of RAM.}. All runs end when the maximum number of generation $G$ is reached, or when the standard deviation of the best fitness over a window of size $L$ reaches some user-defined threshold $\tau_{\sigma}$, whichever comes first.
The code to replicate the comparison experiments will be provided together with the final paper. A file containing the results for all the algorithms, runs and datasets is provided in CSV format in the supplementary material.

\subsection{Hyperparameters} \label{seq:hyperparameters}
ZGP has quite a large number of hyperparameters. On the one hand, this allows a great flexibility when tuning the algorithm. But on the other hand, it makes its use time-consuming. Hence some default values have been fixed by intensive trial-and-error experiments, which are reported in Table~\ref{tab:parameters}.
{The optimization of these hyperparamters by some automatic Hyper Parameter Optimization (HPO) procedure, like SMAC \cite{hutter_sequential_2011}, AutoSkLearn \cite{AutoSkLearn2015} or HyperBand~\cite{li_hyperband_2018} will be the subject of further work.}

\begin{table}[hbt]
\caption{Hyperparameters and default values used in ZGP}
\label{tab:parameters}
\begin{minipage}{\columnwidth}
\begin{center}
\begin{tabular}{p{3cm}l p{2.6cm}}
\toprule
\textbf{Hyperparameter name} & \textbf{Symbol} & \textbf{Value} \\
\midrule
Operator set & $\mathcal{O}$                                                              & \{+,-,*,/,abs, sqrt, sin, cos,  $\lfloor \rfloor$, $\lceil \rceil$, int, mod\}   \\
\# elements   & $n_e$    & 7  \\
\# maturation stages   & $n_m$   & 3   \\
Interval for constants & $[C_{min}$, $C_{max}]$   & [-3, 3]  \\
{Proba. of constants} & {$\rho_{cst}$} & 0.1 \\
Xover tournament size    & $n_{t}$   & 12   \\
{Xover param.} & $\rho_X$ & 0.1 \\
Mutation param. & $m_{mut}$   & 4   \\
Threshold mut. regime   & $r_{lim}$   & 0.1   \\
Population size  & $P$  & 500    \\
Max. \# of generations   & $G$   & 100 \\
Stopping criterion$*$  & $L$, $\tau_{\sigma}$ & 30, $1e^{-3}$\\
\bottomrule
\end{tabular}
\end{center}
\smallskip
\footnotesize $^*$The algorithm is stopped if either one of the two following criterion is reached: $g=G$ (the number of generations reaches the maximum) or the standard deviation of the best fitness over $L$ generations goes below a threshold $\tau_{\sigma}$ (inspired by \cite{rudenko2004steady}).
\end{minipage}
\end{table}

\subsection{Results and Discussion}
As in \cite{orzechowski2018we}, we report the median values for R2 and NRMSE over the 20 runs, for each algorithm and each dataset. 

\begin{figure}[tb]
\begin{subfigure}[b]{\linewidth}
    \centering
    \caption{R2}
    \label{fig:r2-boxplot}
    \includegraphics[width=\linewidth]{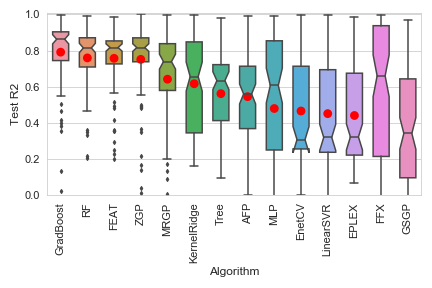}
\end{subfigure}
\newline
\begin{subfigure}[b]{\linewidth}
    \centering
    \caption{NRMSE}
    \label{fig:nrmse-boxplot}
    \includegraphics[width=\linewidth]{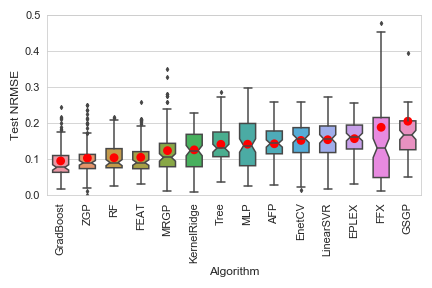}
\end{subfigure}
\caption{Distribution of the median performance (top: R2, bottom: NRMSE) on test set for each dataset. Red points show the average of median R2 over all datasets, and the algorithms are ordered by this measure (best is left, with value closest to 1).}
\label{fig:perf-boxplot}
\end{figure}

Figure~\ref{fig:perf-boxplot} compares the distribution of these median R2 scores (\ref{fig:r2-boxplot}) and NRMSE (\ref{fig:nrmse-boxplot}) over all datasets, while the red dots show the average of the median R2/NRMSE scores over all datasets ("average R2/NRMSE" in the sequel) . The algorithms are ordered by decreasing average R2 and increasing average NRMSE, the best one being on the left. Table~\ref{tab:r2-rank} gives the average rank for each algorithm, based on the median R2 scores (middle column) and median NRMSE (right column) for each dataset. Standard deviations of the ranks are given in parenthesis.

Figure~\ref{fig:perf-boxplot} and Table~\ref{tab:r2-rank}  show that gradient boosting attains the best performance, closely followed by random forests, FEAT and ZGP, and less closely by MRGP. Note that ZGP has lower average R2 than FEAT and better average rank in R2 and NRMSE. This fact comes from a few worse estimations for ZGP on some datasets (lower outliers in R2, Figure~\ref{fig:r2-boxplot}), and better ones for other datasets (higher first and third quartiles in R2). The remaining algorithms display lower performance with a much higher variance, especially on the R2 scores.

\begin{table}[tb]
\caption{Average (and standard deviation) of the ranks in median R2-scores and NRMSE on test set. }
    \label{tab:r2-rank}
    \centering
\begin{tabular}{lp{1.5cm}p{1.5cm}}
\toprule
   Algorithm & R2 avg rank (std) & NRMSE avg rank (std) \\
\midrule
   GradBoost &         3.7 (2.9) &            3.7 (2.9) \\
         \textbf{ZGP} &         \textbf{4.9 (3.0)} &            \textbf{5.0 (3.0)} \\
          RF &         5.0 (2.7) &            5.1 (2.7) \\
        FEAT &         5.6 (2.7) &            5.4 (2.8) \\
 KernelRidge &         6.0 (3.5) &            6.1 (3.5) \\
        MRGP &         7.2 (3.4) &            7.2 (3.3) \\
         FFX &         7.5 (5.1) &            7.5 (5.2) \\
         MLP &         7.9 (4.3) &            7.9 (4.2) \\
        AFP &         8.4 (2.0) &            8.4 (1.9) \\
      EnetCV &         8.4 (4.0) &            8.4 (4.0) \\
   LinearSVR &         9.2 (3.9) &            9.1 (4.1) \\
        Tree &         9.6 (2.9) &            9.6 (3.0) \\
      EPLEX &        10.3 (3.4) &           10.2 (3.4) \\
        GSGP &        11.5 (2.8) &           11.5 (2.8) \\
\bottomrule
\end{tabular}
\end{table}

\begin{figure}[hbt!]
    \centering
\begin{subfigure}[b]{\linewidth}
    \centering
    \caption{All SR algorithms}
    \label{fig:r2-vs-time-all}
    \includegraphics[width=.9\linewidth]{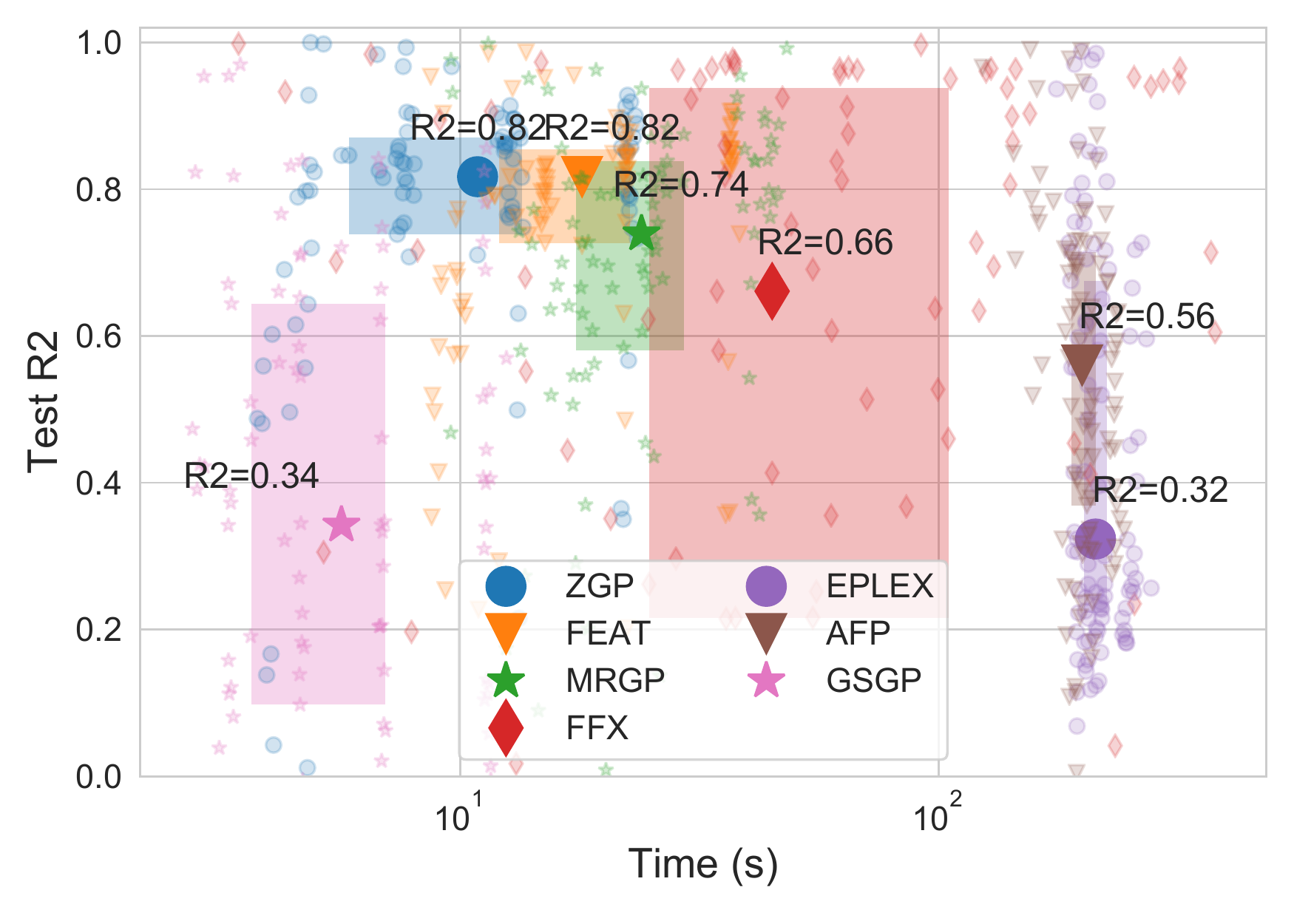}
\end{subfigure}
\newline
\begin{subfigure}[b]{\linewidth}
    \centering
    \caption{Top 3 SR algorithms}
    \label{fig:r2-vs-time-best}
    \includegraphics[width=.9\linewidth]{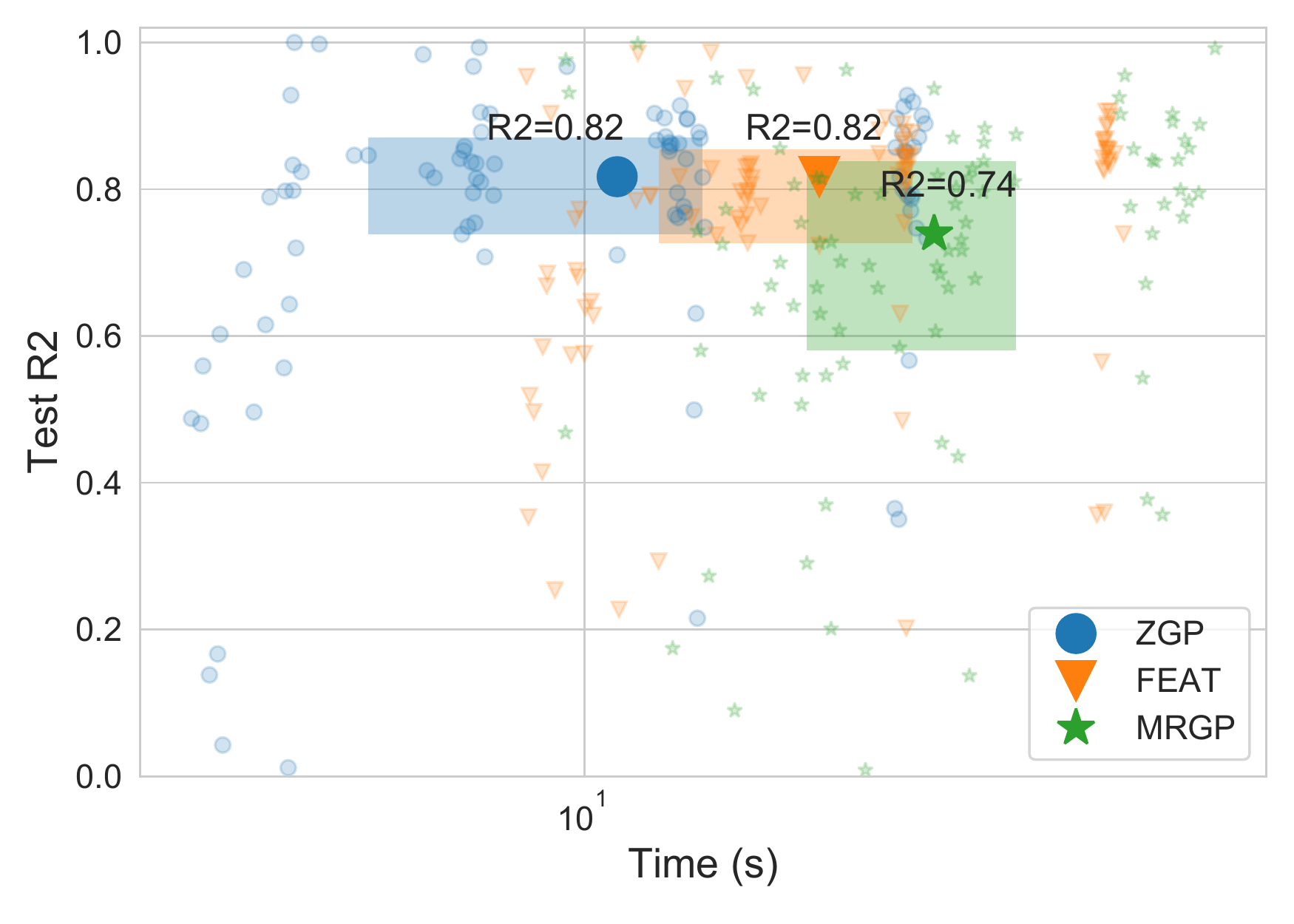}
\end{subfigure}
    \caption{Performance (R2 on test) vs CPU time for SR algorithms (top:all, bottom: best ones, \textcolor{blue}{ZGP}, \textcolor{orange}{FEAT}, \textcolor{ForestGreen}{MRGP}). Transparent markers are the medians for each dataset, large plain markers are the medians over all datasets, and rectangle transparent patches are the 25\%-75\% percentiles. The closest from the upper left corner (high R2, low CPU time) the better. }
    \label{fig:r2-vs-time}
\end{figure}

To further the comparison for symbolic regression, Figure~\ref{fig:r2-vs-time} displays the performance in R2 against the computational time for all SR algorithms (top), and for the top three SR algorithms (bottom), namely ZGP, FEAT and MRGP. A 'good' algorithm should be in the upper left corner of these graphs (high performance and low computational time).
Note that this comparison of runtime is somewhat qualitative because the algorithms rely on different programming languages (ZGP and FEAT are in C++ with a Python interface, while MRGP is in Java). In order to make the comparison as fair as possible, we provided FEAT and MRGP with the maximum execution time as run by ZGP, because both FEAT and MRGP require an upper limit in their input time parameter.

These figures show that ZGP and FEAT share the best performance. Moreover, while all SR algorithms have scattered computational time, GSGP is the fastest SR algorithm, complying with its claim. However, it has a large variance in performance, as does FFX. Among the best three, ZGP thus shows the highest performance and the shortest computational time. 
Also, we note that the algorithms with the lowest performance, GSGP, AFP and EPLEX, are those relying on the smallest operator set $\mathcal{O}=\{+,-,*,/\}$. On the contrary, the best performance is obtained by algorithms with a wider operator set, including trigonometric functions and square roots among others (the full list of operators for each algorithm is provided in supplementary material). {The choice of the operator set appears to be an important one: we investigated it further by expanding it for EPLEX and AFP to $\{+,-,*,/,\sin,\cos,\text{sqrt}\}$, and their performance was indeed greatly improved, but still far from those of ZGP and FEAT; we therefore decided to keep the default set in the results presented here to be consistent with the benchmark in \cite{orzechowski2018we}, and to report the corresponding metrics in supplementary material. Note also that EPLEX and AFP are selection methods, and not full GPSR algorithm, and that EPLEX is used as a selection mechanism for FEAT.}
As for the deterministic FFX, it turns out that both performance and computational time are widely scattered, ranging from very good R2 and low computational time for some datasets, to the worst R2 or computational time on others. It is also worth noting that FFX is the only algorithm that cannot be parametrized directly to run on one thread only and it actually spans all 40 cores of our server, while all the other algorithms were limited to one core per run. 

We are aware of the limitations of the datasets utilized. Despite their number, as mentioned in this section and in the introduction, a good portion (62 of them) consists of simulated data from the Friedman collection of artificial datasets, that follow a known nonlinear function with only 3 relevant variables, as described in \cite{friedman2001greedy}. 
Hence, the chosen database may be a favorable setting for symbolic regression approaches that tend to select few variables in the final model (among which ZGP).


\section{Conclusion and future work} \label{sec:discussion}
ZGP, a novel GPSR algorithm, is presented and its inner working  explained in detail. The algorithm has been validated on regression tasks, in comparison with several-state-of-the art algorithms, both from classic ML tools and from existing GP-based SR frameworks.

The performance of ZGP is comparable or better than the state-of-the-art SR algorithms, in terms of accuracy of the resulting model (measured both by the RMSE or the R2), and of computational time. It is also comparable to state-of-the-art classic ML algorithms, but performs somewhat worse than the most advanced ML algorithms like gradient boosting. However, the comparison of the computational time is semi-quantitative as it is difficult to guarantee conditions that are fully equivalent for all.

Similarly to the majority of the SR algorithms, ZGP's interpretability is attained through a tight selection of variables, and the output of an analytical formula, linking the selected variables to the target. However, and different from most other SR algorithms, ZGP is "bloat-adverse by design": the zoetropic representation, and the genotype-to-phenotype mapping give an upper bound for the complexity of all ZGP models. Last but not least, ZGP performs feature construction and selection: the "zoetropes", the elements obtained on the last layer of the development process, are simply combined by linear regression. Therefore, they do represent useful features, being a by-product of the  algorithm rather than a separate pre-processing step. These features offer yet another insight on the interpretation of the model.

One of the specifics of ZGP is the use of the fusion operation during the genotype-to-phenotype mapping (see Section \ref{sec:z-forge} and in particular Equation \ref{eq:fusion}). No operator is applied alone, and smooth transitions from one operator to the other are possible through modifications by mutation of the random weight $r$, in a way similar to that of Geometric Semantic Crossover \cite{moraglio2012geometric}. However, the linear combination is limited here to simple operators, and is only performed $n_m$ times, thus does not result in uncontrolled bloat: instead of increasing the search space by augmenting the complexity of the trees, as in traditional GP, the search space is extended in ZGP by replacing the discrete set of operators by the continuous family obtained by their linear combinations. On-going ablation studies are investigating this hypothesis.

In contrast to algorithms designed for big data, ZGP, like all GP-based SR algorithms, can attain its results by handling datasets with less than a few thousands of observations (less than 3000 in the present experiments), a context often loosely referred to today  as "small data". {Whereas emphasis has been put on Big Data in the recent years due to impressive results in image recognition and Natural Language Processing, to name a few, for many more companies out there the available data does not qualify as "Big".} \\

As mentioned in the introduction, ZGP can also be applied to classification and benchmarking in both binary, and multi-class tasks is the subject of on-going work.
Furthermore, an extension of the benchmark to a database including more real-world datasets for all three tasks will provide a fuller assessment for those algorithms. 
Even though the inner mechanism of the ZGP algorithm does limit the bloat, a major effort for future work is the quantitative assessment of the model complexity. 
Several measures may capture the complexity of symbolic regression models, and we plan to assess it as an additional level for model selection.
Finally, hyperparameter tuning is often performed ad-hoc, whereas a systematic treatment 
may help, in particular to select the number of elements and stages, which can be constraining at present.

\small
\bibliographystyle{abbrv}

\begin{thebibliography}{10}

\bibitem{affenzeller2019white}
M.~Affenzeller, B.~Burlacu, V.~Dorfer, S.~Dorl, G.~Halmerbauer,
  T.~K{\"o}nigswieser, M.~Kommenda, J.~Vetter, and S.~Winkler.
\newblock White box vs. black box modeling: On the performance of deep
  learning, random forests, and symbolic regression in solving regression
  problems.
\newblock In {\em International Conference on Computer Aided Systems Theory},
  pages 288--295. Springer, 2019.

\bibitem{arnaldo2014multiple}
I.~Arnaldo, K.~Krawiec, and U.-M. O'Reilly.
\newblock Multiple regression genetic programming.
\newblock In {\em Proceedings of the 2014 Annual Conference on Genetic and
  Evolutionary Computation}, pages 879--886, 2014.

\bibitem{arnaldo2015building}
I.~Arnaldo, U.-M. O'Reilly, and K.~Veeramachaneni.
\newblock Building predictive models via feature synthesis.
\newblock In {\em Proceedings of the 2015 Annual Conference on Genetic and
  Evolutionary Computation}, pages 983--990, 2015.

\bibitem{banzhaf98}
W.~Banzhaf, P.~Nordin, R.~E. Keller, and F.~D. Francone.
\newblock {\em Genetic programming}.
\newblock Springer, 1998.

\bibitem{bishop2006pattern}
C.~M. Bishop.
\newblock {\em {Pattern recognition and machine learning}}.
\newblock Information science and statistics. Springer, New York, NY, 2006.
\newblock Softcover published in 2016.

\bibitem{breiman2001random}
L.~Breiman.
\newblock Random forests.
\newblock {\em Machine learning}, 45(1):5--32, 2001.

\bibitem{castelli2019gsgp}
M.~Castelli and L.~Manzoni.
\newblock Gsgp-c++ 2.0: A geometric semantic genetic programming framework.
\newblock {\em SoftwareX}, 10:100313, 2019.

\bibitem{chollet2015keras}
F.~Chollet et~al.
\newblock Keras.
\newblock \url{https://keras.io}, 2015.

\bibitem{defranca2020interaction}
F.~O. de~Fran{\c{c}}a and G.~S.~I. Aldeia.
\newblock Interaction-transformation evolutionary algorithm for symbolic
  regression.
\newblock {\em Evolutionary Computation}, pages 1--25, 2020.

\bibitem{AutoSkLearn2015}
M.~Feurer, A.~Klein, K.~Eggensperger, J.~Springenberg, M.~Blum, and F.~Hutter.
\newblock Efficient and robust automated machine learning.
\newblock In C.~Cortes, N.~Lawrence, D.~Lee, M.~Sugiyama, and R.~Garnett,
  editors, {\em Advances in Neural Information Processing Systems}, volume~28,
  pages 2962--2970. Curran Associates, Inc., 2015.

\bibitem{friedman2001greedy}
J.~H. Friedman.
\newblock Greedy function approximation: a gradient boosting machine.
\newblock {\em Annals of statistics}, pages 1189--1232, 2001.

\bibitem{hastie2009elements}
T.~Hastie, R.~Tibshirani, and J.~Friedman.
\newblock {\em The elements of statistical learning: data mining, inference,
  and prediction}.
\newblock Springer Science \& Business Media, 2009.

\bibitem{hutter_sequential_2011}
F.~Hutter, H.~H. Hoos, and K.~Leyton-Brown.
\newblock Sequential {Model}-based {Optimization} for {General} {Algorithm}
  {Configuration}.
\newblock In {\em Proceedings of the 5th {International} {Conference} on
  {Learning} and {Intelligent} {Optimization}}, {LION}'05, pages 507--523,
  Berlin, Heidelberg, 2011. Springer-Verlag.
\newblock event-place: Rome, Italy.

\bibitem{kim2020integration}
S.~Kim, P.~Y. Lu, S.~Mukherjee, M.~Gilbert, L.~Jing, V.~{\v{C}}eperi{\'c}, and
  M.~Solja{\v{c}}i{\'c}.
\newblock Integration of neural network-based symbolic regression in deep
  learning for scientific discovery.
\newblock {\em IEEE Transactions on Neural Networks and Learning Systems},
  2020.

\bibitem{korns2011accuracy}
M.~F. Korns.
\newblock Accuracy in symbolic regression.
\newblock In {\em Genetic Programming Theory and Practice IX}, pages 129--151.
  Springer, 2011.

\bibitem{Koza92}
J.~R. Koza.
\newblock {\em Genetic programming: on the programming of computers by means of
  natural selection}, volume~1.
\newblock MIT press, 1992.

\bibitem{lacava2015genetic}
W.~La~Cava, T.~Helmuth, L.~Spector, and K.~Danai.
\newblock Genetic programming with epigenetic local search.
\newblock In {\em Proceedings of the 2015 Annual Conference on Genetic and
  Evolutionary Computation}, pages 1055--1062, 2015.

\bibitem{lacava2019learning}
W.~La~Cava, T.~R. Singh, J.~Taggart, S.~Suri, and J.~H. Moore.
\newblock Learning concise representations for regression by evolving networks
  of trees.
\newblock In {\em International {Conference} on {Learning} {Representations}},
  {ICLR}, 2019.

\bibitem{lacava2016epsilon}
W.~La~Cava, L.~Spector, and K.~Danai.
\newblock Epsilon-lexicase selection for regression.
\newblock In {\em Proceedings of the Genetic and Evolutionary Computation
  Conference 2016}, pages 741--748, 2016.

\bibitem{langdon2013foundations}
W.~B. Langdon and R.~Poli.
\newblock {\em Foundations of genetic programming}.
\newblock Springer Science \& Business Media, 2013.

\bibitem{li_hyperband_2018}
L.~Li, K.~Jamieson, G.~DeSalvo, A.~Rostamizadeh, and A.~Talwalkar.
\newblock Hyperband: {A} {Novel} {Bandit}-{Based} {Approach} to
  {Hyperparameter} {Optimization}.
\newblock {\em Journal of Machine Learning Research}, 18(185):1--52, 2018.

\bibitem{mcconaghy2011ffx}
T.~McConaghy.
\newblock {FFX}: Fast, scalable, deterministic symbolic regression technology.
\newblock In {\em Genetic Programming Theory and Practice IX}, pages 235--260.
  Springer, 2011.

\bibitem{mcdermott2012genetic}
J.~McDermott, D.~R. White, S.~Luke, L.~Manzoni, M.~Castelli, L.~Vanneschi,
  W.~Jaskowski, K.~Krawiec, R.~Harper, K.~De~Jong, et~al.
\newblock Genetic programming needs better benchmarks.
\newblock In {\em Proceedings of the 14th annual conference on Genetic and
  evolutionary computation}, pages 791--798, 2012.

\bibitem{moraglio2012geometric}
A.~Moraglio, K.~Krawiec, and C.~G. Johnson.
\newblock Geometric semantic genetic programming.
\newblock In {\em International Conference on Parallel Problem Solving from
  Nature}, pages 21--31. Springer, 2012.

\bibitem{olson2017pmlb}
R.~S. Olson, W.~La~Cava, P.~Orzechowski, R.~J. Urbanowicz, and J.~H. Moore.
\newblock {PMLB}: a large benchmark suite for machine learning evaluation and
  comparison.
\newblock {\em BioData Mining}, 10(1):36, Dec 2017.

\bibitem{orzechowski2018we}
P.~Orzechowski, W.~La~Cava, and J.~H. Moore.
\newblock Where are we now? a large benchmark study of recent symbolic
  regression methods.
\newblock In {\em Proceedings of the Genetic and Evolutionary Computation
  Conference}, pages 1183--1190, 2018.

\bibitem{scikit-learn}
F.~Pedregosa, G.~Varoquaux, A.~Gramfort, V.~Michel, B.~Thirion, O.~Grisel,
  M.~Blondel, P.~Prettenhofer, R.~Weiss, V.~Dubourg, J.~Vanderplas, A.~Passos,
  D.~Cournapeau, M.~Brucher, M.~Perrot, and E.~Duchesnay.
\newblock Scikit-learn: Machine learning in {P}ython.
\newblock {\em Journal of Machine Learning Research}, 12:2825--2830, 2011.

\bibitem{poli2008field}
R.~Poli, W.~B. Langdon, N.~F. McPhee, and J.~R. Koza.
\newblock {\em A field guide to genetic programming}.
\newblock Lulu. com, 2008.

\bibitem{ribeiro2016should}
M.~T. Ribeiro, S.~Singh, and C.~Guestrin.
\newblock " why should i trust you?" explaining the predictions of any
  classifier.
\newblock In {\em Proceedings of the 22nd ACM SIGKDD international conference
  on knowledge discovery and data mining}, pages 1135--1144, 2016.

\bibitem{rudenko2004steady}
O.~Rudenko and M.~Schoenauer.
\newblock A steady performance stopping criterion for pareto-based evolutionary
  algorithms.
\newblock In {\em 6th International Multi-Objective Programming and Goal
  Programming Conference}, 2004.

\bibitem{schmidt2011age}
M.~Schmidt and H.~Lipson.
\newblock Age-fitness {Pareto} optimization.
\newblock In {\em Genetic programming theory and practice VIII}, pages
  129--146. Springer, 2011.

\bibitem{tao1998inver}
G.~Tao and Z.~Michalewicz.
\newblock Inver-over operator for the tsp.
\newblock In {\em International Conference on Parallel Problem Solving from
  Nature}, pages 803--812. Springer, 1998.

\bibitem{tibshirani2011regression}
R.~Tibshirani.
\newblock Regression shrinkage and selection via the lasso: a retrospective.
\newblock {\em Journal of the Royal Statistical Society: Series B (Statistical
  Methodology)}, 73(3):273--282, 2011.

\bibitem{vanneschi2014survey}
L.~Vanneschi, M.~Castelli, and S.~Silva.
\newblock A survey of semantic methods in genetic programming.
\newblock {\em Genetic Programming and Evolvable Machines}, 15(2):195--214,
  2014.

\bibitem{virgolin2020improving}
M.~Virgolin, T.~Alderliesten, C.~Witteveen, and P.~A.~N. Bosman.
\newblock Improving model-based genetic programming for symbolic regression of
  small expressions.
\newblock {\em Evolutionary Computation}, page 1–27, Jun 2020.

\bibitem{vries2018sensitivity}
S.~d. Vries.
\newblock Sensitivity analysis based feature-guided evolution for symbolic
  regression.
\newblock Master's thesis, 2018.

\bibitem{vzegklitz2017symbolic}
J.~{\v{Z}}egklitz and P.~Po{\v{s}}{\'\i}k.
\newblock Symbolic regression algorithms with built-in linear regression.
\newblock {\em arXiv preprint arXiv:1701.03641}, 2017.

\bibitem{vzegklitz2020benchmarking}
J.~{\v{Z}}egklitz and P.~Po{\v{s}}{\'\i}k.
\newblock Benchmarking state-of-the-art symbolic regression algorithms.
\newblock {\em Genetic Programming and Evolvable Machines}, pages 1--29, 2020.

\bibitem{zou2005regularization}
H.~Zou and T.~Hastie.
\newblock Regularization and variable selection via the elastic net.
\newblock {\em Journal of the royal statistical society: series B (statistical
  methodology)}, 67(2):301--320, 2005.

\end{thebibliography}

\end{document}